\documentclass[10pt, logo, twocolumn, copyright]{nvidiatechreport}

\usepackage{mdframed}
\usepackage[utf8]{inputenc} 
\usepackage[T1]{fontenc}    

\usepackage{amsfonts}       
\usepackage{nicefrac}       
\usepackage{microtype}      
\usepackage[dvipsnames]{xcolor}         
\usepackage{multirow}
\usepackage{multicol}
\usepackage{graphicx}
\usepackage[numbers]{natbib}
\usepackage{tabto}
\usepackage{xspace}
\usepackage{amsmath}
\usepackage{adjustbox}
\usepackage{enumitem}
\usepackage{wrapfig}
\usepackage{dblfloatfix}
\usepackage{tcolorbox}
\usepackage{listings}
\usepackage{footmisc}

\usepackage{float}

\usepackage{natbib}

\usepackage{hyperref}
\usepackage{url}
\usepackage{booktabs}
\usepackage{multirow}
\usepackage{enumitem}
\usepackage{adjustbox}
\usepackage{tabularx}
\usepackage{arydshln}
\usepackage{wrapfig}

\usepackage{multirow}
\usepackage{colortbl}
\usepackage{amsmath}
\usepackage{makecell}
\usepackage{hhline}
\usepackage{array}
\usepackage{diagbox}
\usepackage{graphicx}
\usepackage{adjustbox}
\usepackage{bm}

\usepackage{fp} 

\usepackage{xcolor}
\definecolor{mygreen}{rgb}{0.7, 1.0, 0.7}

\definecolor{myblue}{rgb}{0.7, 0.8, 1.0}

\usepackage{pifont}
\newcommand\mynuma[1]{\ifcase#1 \or \ding{172}\or \ding{173}\or
  \ding{174}\or \ding{175}\or \ding{176}\or \ding{177}%
  \or \ding{178}\or \ding{179}\or \ding{180}\or \ding{181}\else *\fi\relax}

\newcommand\mynumb[1]{\ifcase#1 \or \ding{182}\or \ding{183}\or
  \ding{184}\or \ding{185}\or \ding{186}\or \ding{187}%
  \or \ding{188}\or \ding{189}\or \ding{190}\or \ding{191}\else *\fi\relax}

\newcounter{mycounter} 

\title{C-RADIOv4 (Tech Report)}

\author{Mike Ranzinger$^*$ \qquad Greg Heinrich$^*$ \qquad Collin McCarthy \qquad Jan Kautz \qquad
Andrew Tao \qquad Bryan Catanzaro \qquad Pavlo Molchanov
}

\begin{document}

\begin{abstract}
By leveraging multi-teacher distillation, agglomerative vision backbones provide a unified student model that retains and improves the distinct capabilities of multiple teachers. In this tech report, we describe the most recent release of the C-RADIO family of models, C-RADIOv4, which builds upon AM-RADIO/RADIOv2.5 in design, offering strong improvements on key downstream tasks at the same computational complexity. We release -SO400M (412M params), and -H (631M) model variants, both trained with an updated set of teachers: SigLIP2, DINOv3, and SAM3. In addition to improvements on core metrics and new capabilities from imitating SAM3, the C-RADIOv4 model family further improves any-resolution support, brings back the ViTDet option for drastically enhanced efficiency at high-resolution, and comes with a permissive license.

\vspace{5pt}

\textbf{Links:} \hspace{2pt} \href{https://github.com/NVlabs/RADIO}{Code} (on GitHub) | \href{https://huggingface.co/collections/nvidia/radio-669f77f1dd6b153f007dd1c6}{Models} (on Hugging Face) 

\vspace{10pt}

\end{abstract} 
\maketitle
\renewcommand{\thefootnote}{\fnsymbol{footnote}}
\footnotetext[1]{Equal Contribution}
\renewcommand{\thefootnote}{\arabic{footnote}}
\section{Description}
\label{sec:description}
AM-RADIO \cite{ranzinger2024radio} introduced the concept of agglomerative foundation models, which is a method of creating a new foundation model by distilling the feature representations from other heterogeneous models. In the original formulation, we used DFN CLIP \cite{fang2023datafilteringnetworks}, DINOv2 \cite{darcet2024visiontransformersneedregisters}, and SAM \cite{kirillov2023sam} as the core teacher set. We also introduced multi-resolution training in the sense that DFN CLIP and DINOv2 were distilled at one resolution, and SAM was distilled at a higher resolution. We identified that there was a phenomenon we called ``mode switching'' where the student model learned to change its representations based on the resolution in order to minimize the training loss, and it caused very inconsistent behavior during inference depending on the resolution used. In PHI-S \cite{ranzinger2024phisdistributionbalancinglabelfree} we identified teacher distribution balancing as an important component to training strong agglomerative models, in RADIOv2.5 \cite{Heinrich_2025_CVPR} we found that training against all teachers at both resolutions was sufficient to overcome the mode switching issue, and in FeatSharp \cite{ranzinger2025featsharp} we proposed an upsampling method which is suitable for certain fixed-resolution models such as DFN CLIP and SigLIP \cite{zhai2023sigmoidlosslanguageimage,tschannen2025siglip2multilingualvisionlanguage}, which is preferable compared to bilinear resampling. One of the key points made in the original AM-RADIO was that improved teachers tend to yield improved students, and this trend continues to hold. Since the previous works, SigLIP2 \cite{tschannen2025siglip2multilingualvisionlanguage} has become the frontier text-image foundation encoder, DINOv3 \cite{simeoni2025dinov3} has pushed the boundaries of self-supervised learning (SSL) and dense representations, yielding an incredibly strong dense perception model, and SAM has upgraded to SAM3 \cite{carion2025sam3segmentconcepts} making large inroads toward solving computer vision. In keeping with our initial premise, we have upgraded the core set of teachers to [SigLIP2, DINOv3, SAM3], to upgrade our agglomerative model. We inherit DINOv3's improved semantic segmentation capability, and SigLIP2's enhanced text alignment. SAM3 as a teacher doesn't show improved metrics on our selected benchmarks, but including it as a teacher allows for replacing the vision encoder backbone of SAM3, along with creative use cases which exploit agglomerative models, such as that in RADSeg \cite{alama2025radsegunleashingparametercompute}. Beyond simply updating our teacher set, we further our commitment to versatility by improving the ability of our model to operate at any resolution, and bring back ``ViTDet mode'', allowing most of the transformer blocks to operate in windowed mode, which has a dramatic effect on inference speed on high-resolution images (see figure \ref{fig:sam3_benchmark_results}).
\begin{table*}[]
    \centering
    \begin{tabular}{lcr|cc|cc|cc}
        \multicolumn{3}{c|}{\bf{Model}} & \multicolumn{2}{|c|}{\bf{Summary}} & \multicolumn{2}{|c|}{\bf{Segmentation}} & \multicolumn{2}{|c}{\bf{Probe3d}} \\
        Name & Variant & Params & Zero Shot & kNN & ADE20k & VOC & NAVI & SPair \\
        \hline
        \multirow{2}{*}{SigLIP2} & SO400M & 412M & \underline{83.88} & 85.76 & 44.0 & & 48.8 & 38.7 \\
                                 & g & 1,164M & \bf{84.75} & \underline{86.39} & 42.7 & 72.7 & 49.40 & 42.60 \\
        \hline
        \multirow{2}{*}{DINOv3} & H+ & 841M & - & 85.77 & 54.8 & & 63.3 & 56.3 \\
                                & 7B & 6,716M   & - & 85.42 & \bf{55.9} & 86.6 & \bf{64.4} & 58.7 \\
        \hline
        RADIOv2.5 & H & 631M & 82.51 & 85.81 & 51.58 & 85.97 & 60.89 & 56.24 \\
        C-RADIOv3 & H & 631M & 82.65 & 86.23 & 52.75 & 86.41 & 62.10 & 58.54 \\
        \hline
        \multirow{2}{*}{C-RADIOv4} & SO400M & 412M & 82.01 & 85.76 & 55.14 & \underline{87.22} & 62.44 & 60.01 \\
                                   & H & 631M & 83.09 & \bf{86.59} & \underline{55.20} & \bf{87.24} & \underline{63.44} & \bf{60.57} \\
    \end{tabular}
    \caption{Comparison between different vision encoders. C-RADIOv4 is competitive with DINOv3 on dense tasks at a fraction of the parameters.}
    \label{tab:model_comparison}
\end{table*}

\begin{figure*}[!t]
    \centering
    \includegraphics[width=\linewidth]{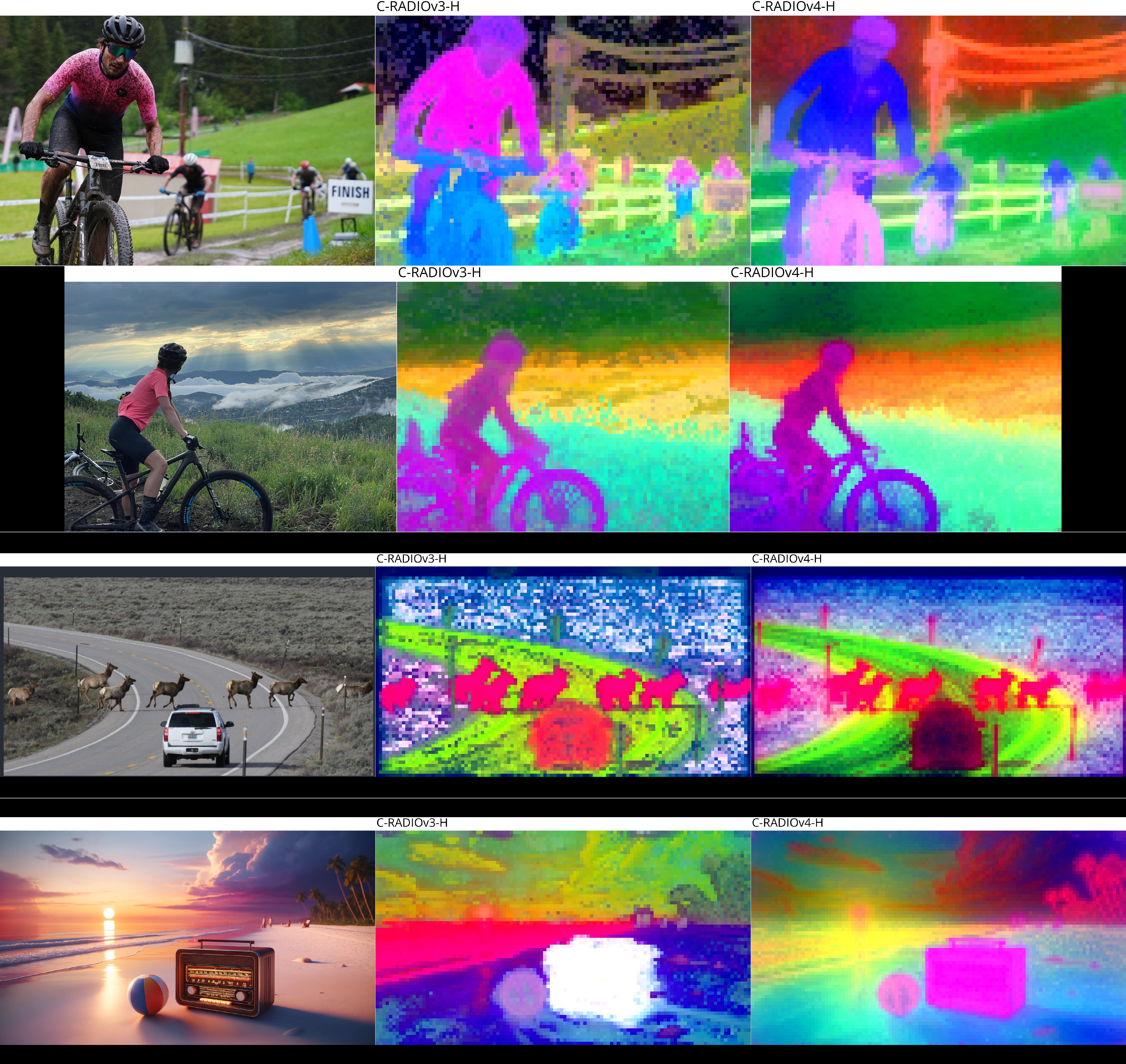}
    \caption{PCA feature visualization, comparing C-RADIOv3-H vs C-RADIOv4-H. Object boundaries are now significantly cleaner.}
    \label{fig:pca_viz}
\end{figure*}

\section{Method Updates}

Agglomerative models rely on distillation from multiple vision foundation models. For the RADIO family of models, the distillation operates over both the dense feature space and summarization tokens. In this section, we describe the novel techniques employed to develop the improved C-RADIOv4 model over previous variants in the form of what has changed since we fully described the method in RADIOv2.5 \cite{Heinrich_2025_CVPR}.

\subsection{Updated Teachers}

For C-RADIOv4, we update the teacher set to:
\begin{samepage}
\begin{itemize}
    \item SigLIP2-g-384 \cite{tschannen2025siglip2multilingualvisionlanguage}
    \item DINOv3-7B \cite{simeoni2025dinov3}
    \item SAM3 \cite{carion2025sam3segmentconcepts}
\end{itemize}
\end{samepage}

In order to reduce the computational demand, we dropped support for DFN CLIP \cite{fang2023datafilteringnetworks} in this release, in favor of the more ubiquitous SigLIP2 \cite{tschannen2025siglip2multilingualvisionlanguage}, as the latter is being deployed in more places (e.g. Qwen3 VL \cite{bai2025qwen3vltechnicalreport}), and because both models have similar representations and application domains.

\subsection{Stochastic Resolutions}

Instead of training at 2 different resolutions as in RADIOv2.5 \cite{Heinrich_2025_CVPR}, we sample from $\{128, 192, 224, 256, 384, 432\}$ in the low-resolution partition, and $\{512, 768, 1024, 1152\}$ in the high-resolution partition, enabling an even smoother resolution scaling curve, and also gaining substantial quality at low resolutions. For SigLIP2 \cite{tschannen2025siglip2multilingualvisionlanguage}, we use FeatSharp \cite{ranzinger2025featsharp} to do $3 \times$ upsampling from 384px to 1152px in our high-resolution training partition. The low-resolution partition uses the raw outputs. Finally, for SAM3, we use the mosaic augmentation as proposed in RADIOv2.5 \cite{Heinrich_2025_CVPR}, since it only supports inputs of size $1152 \times 1152$. Figures \ref{fig:zero_shot_resolution_scaling} and \ref{fig:knn_resolution_scaling} show how multi-resolution support has evolved over time, with figure~\ref{fig:knn_resolution_scaling} showing how RADIO models compare against DINOv2/3, as they also support multi-res. In table \ref{tab:mmseg_resolution_scaling} we show that C-RADIOv4 is strongly robust to increasing the resolution for semantic segmentation, including at resolutions higher than trained. Considering that C-RADIOv4-H has an order of magnitude fewer parameters than DINOv3-7B, it performs very competitively. 

\begin{table}[]
    \centering
    \begin{tabular}{r|ccc}
        \multirow{2}{*}{\bf{Model}} & \multicolumn{3}{c}{\bf{ADE20k}} \\
                                   & 512px      & 1024px            & 1536px  \\
        \hline
        DINOv3-7B                  & 55.9       & 57.3${}^\dagger$ & 57.8${}^\dagger$     \\
        C-RADIOv4-H                & 55.20      & 57.02            & 57.72                \\
    \end{tabular}
    \caption{Comparison of resolution scaling for ADE20k \cite{zhou2018semanticunderstandingscenesade20k} linear probe between DINOv3-7B and C-RADIOv4-H. Both models exhibit strong resolution scaling properties. ${}^\dagger$The metrics for DINOv3-7B come from figure 11 in \cite{simeoni2025dinov3}, where values past the decimal are approximate.}
    \label{tab:mmseg_resolution_scaling}
\end{table}

\subsection{Shift Equivariance}

\begin{figure*}[t]
    \centering
    \includegraphics[width=\linewidth]{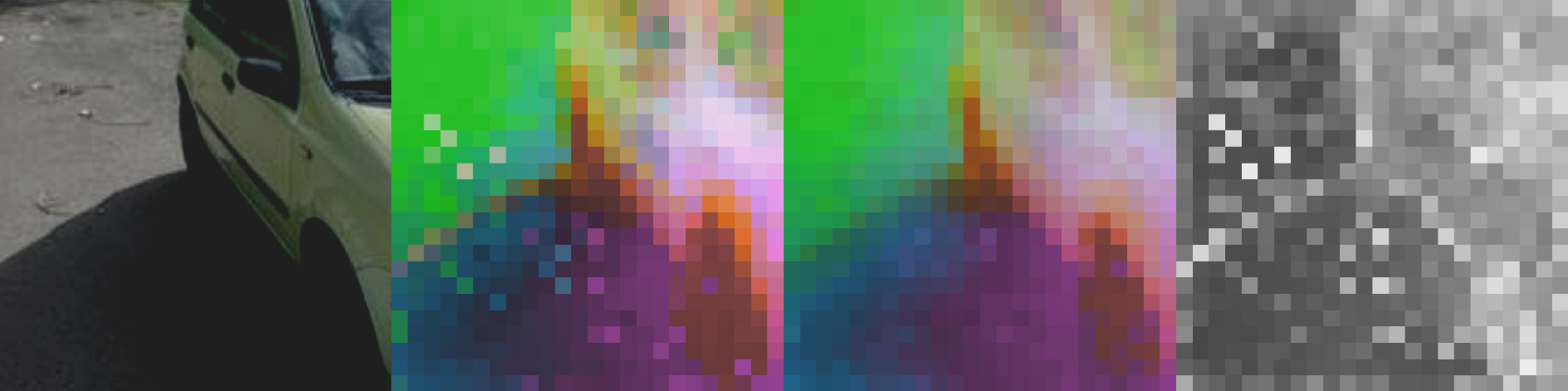}
    \includegraphics[width=\linewidth]{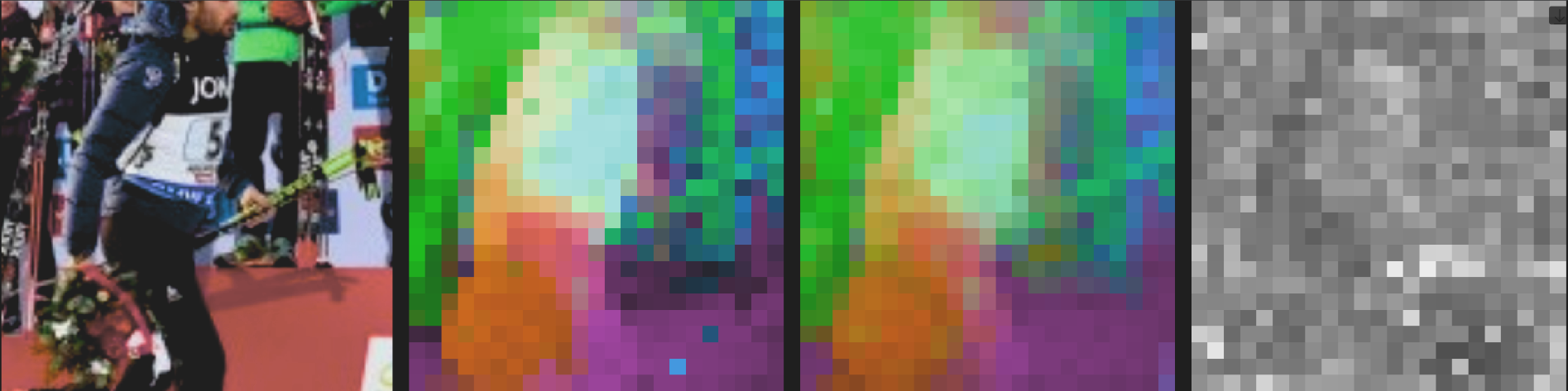}
    \includegraphics[width=\linewidth]{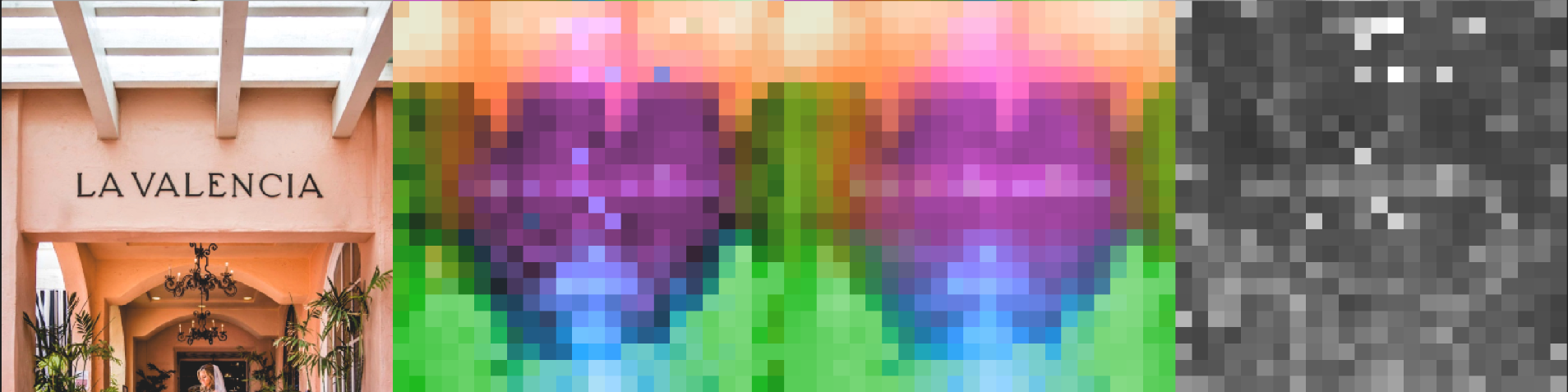}
    \caption{Visualization of DINOv3 and C-RADIOv4 (adapter) predictions. Notice the out-of-place speckles produced by DINOv3. \\
    \textbf{Column 1:} Input Image \\
    \textbf{Column 2:} DINOv3 PCA visualization. \\
    \textbf{Column 3:} C-RADIOv4 adapter PCA visualization. \\
    \textbf{Column 4:} Error heatmap between the student adapter prediction and the DINOv3 teacher.
    }
    \label{fig:dv3_artifacts}
\end{figure*}

Since PHI-S \cite{ranzinger2024phisdistributionbalancinglabelfree}, the formulation of the feature loss has been $\frac{1}{Z} \sum_i (x - \hat{y})^2$ with $\hat{y}$ being the teacher features normalized by PHI-S, and $x$ being the student prediction, and $Z$ the normalizing constant. A drawback of this approach is that the student is not only learning the useful features of the teacher, but also the fixed pattern noise. In DVT \cite{yang2024dvt}, the authors find that all of these vision foundation models have this noise, and they learn to disentangle the models outputs into $ViT(x) \approx f(x)+g(E_{pos})+h(x,E_{pos})$, with $f$ being the input-dependent semantics, $g$ being a data-invariant bias, and $h$ the entangled residual. We also identified these same noise patterns in the FeatSharp \cite{ranzinger2025featsharp} work, notably with the SigLIP2 models, as they have these ``holes'' along the border of the output feature maps. SAM \cite{kirillov2023sam} has strong artifacts at the borders of the ViTDet \cite{li2022vitdet} windows. DINOv3-H+ happens to have somewhat frequent large magnitude noise patches. In all of these cases, left unmitigated, the model will learn to mimic this noise in the MLP adapter, and it will even leak down into the backbone features. We combat this with two different forms of shift equivariance in the loss formulation, both of which make it impossible for the student to know the exact positions of the patches being matched between student and teacher. We visualize these artifacts in figure \ref{fig:dv3_artifacts}, where high-energy representations appear, often in otherwise mostly uniform image regions.

\subsubsection{Shift Equivariant Loss}

To prevent the student from matching dense teacher features which are not explicitly input-dependent semantics \cite{yang2024dvt}, for a particular image, we randomly shift the crop seen by the student, and also each teacher. The teachers get independent shifts relative to each other, and the student. We track a mapping $\mathcal{F}_{S\rightarrow T}$ which transforms the features produced by the student to be spatially aligned with the teacher. The sampled shifts are in increments of the patch size, which eliminates interpolation effects where possible. Thus, we adopt a new loss formulation as follows:

\begin{equation}
    L_{\mathrm{spatial}}(\mathbf{x}, \hat{\mathbf{y}})
    = \frac{1}{|\Omega|} \sum_{u \in \Omega}
    \left( \mathcal{F}_{S\rightarrow T}\!\left[\mathbf{x}\right]_u - \hat{\mathbf{y}}_u \right)^2 .
\end{equation}

With $\Omega$ being the set of common spatial positions seen between student and teacher, $\mathbf{x}$ the student output, and $\hat{\mathbf{y}}$ the PHI-S \cite{ranzinger2024phisdistributionbalancinglabelfree} normalized teacher output.

\subsubsection{Shift Equivariant MESA}

MESA \cite{du2022mesa} is an indispensable tool for training classification models, which attempts to converge the weights to flat regions where perturbations to the input don't cause chaotic changes in the output and improves generalization. To further combat fixed-pattern noise from emerging in our student model, we apply the MESA formulation of matching the exponential moving average (EMA) of the student model, but with the added twist of introducing different crops for the student and its EMA, and relating them via the $\mathcal{F}_{S\rightarrow \tilde{S}}$ transform, and then use the following formulation:

\begin{equation}
    L_{mesa}(\mathbf{x}, \tilde{\mathbf{x}}) = \frac{1}{|\Omega|} \sum_{u \in \Omega} \left(\mathcal{F}_{S\rightarrow \tilde{S}}\left[LN(\mathbf{x})\right]_u - LN(\tilde{\mathbf{x}})_u \right)^2
\end{equation}

\noindent with $LN$ being layer norm, without the learnable affine projection, and $\tilde{S},\tilde{\mathbf{x}}$ the respective crop and output of the EMA student model.

\subsection{DAMP}

To further encourage the robustness of our model, we employ DAMP \cite{trinh2024damp}, which applies multiplicative noise to the weights of our model during training.

\subsection{Balanced Summary Loss}

In PHI-S \cite{ranzinger2024phisdistributionbalancinglabelfree}, the entire study revolved around normalizing the distributions of the spatial features of each teacher, so that large activations wouldn't dominate the loss. However, no attention was paid to the summary features. The reason for this was because the summary features were matched using cosine similarity, which is inherently normalized. However, what we have since noticed is that while the \textit{magnitudes} of the summary features are normalized onto the unit hypersphere, their directional variance is not. If teachers produced features with a uniform distribution over the unit hypersphere, this would not be important, but instead what we found is that features tend to fall into a cone, and the radius of this cone is different for each teacher. The effect of this is that cones with larger radius will produce larger losses than those with small radius. Further, it's not so much the angle between student and teacher that matters, but rather that angle relative to all other embeddings. To mitigate this, we no longer use cosine distance as our summary loss, and instead adopt the following:

\begin{align}
    \cos(\mathbf{x}, \mathbf{y}) &= \frac{\mathbf{x}^\intercal \mathbf{y}}{\|\mathbf{x}\|\|\mathbf{y}\|} \\
    \Theta(\mathbf{x},\mathbf{y}) &= \arccos(\cos(\mathbf{x},\mathbf{y})) \\
    \bm{\mu}_\mathbf{y} &= \frac{\mathbb{E}[\mathbf{y}]}{\|\mathbb{E}[\mathbf{y}]\|} \\
    \text{Disp}(\Theta_\mathbf{y}) &= \mathbb{E}\left[\Theta \left(\mathbf{y}, \bm{\mu}_\mathbf{y} \right)^2 \right] \label{eq:angle_variance} \\
    L_{angle}(\mathbf{x},\mathbf{y}) &= \frac{\Theta(\mathbf{x},\mathbf{y})^2}{\text{Disp}(\Theta_\mathbf{y})}
\end{align}

With $\mathbf{x}$ the student predictions, $\mathbf{y}$ the teacher prediction, $\bm{\mu}_\mathbf{y}$ being the expected direction of $\mathbf{y}$, and $\text{Disp}(\Theta_\mathbf{y})$ being the angular dispersion of $\mathbf{y}$, which we use to normalize the cone radius between teachers, allowing the student to focus on the relative angles, and to prevent one teacher from dominating another. We report the angular dispersions we found in Table \ref{tab:angle_variance}, showing that there is a significant difference between SigLIP2 and DINOv3. Left unmitigated, DINOv3 would dominate the loss term, biasing the student toward matching it, at the expense of matching SigLIP2.

\begin{table}[]
    \centering
    \begin{tabular}{c|c}
        \bf{Model} & $\text{\bf{Disp}}(\Theta_\mathbf{y})$ \\
        \hline
        SigLIP2-g-384 & 0.694 \\
        DINOv3-H+ & 2.120 \\
        DINOv3-7B & 2.186 \\
    \end{tabular}
    \caption{The angular dispersion of the summary token for key teachers.}
    \label{tab:angle_variance}
\end{table}

\section{Results}

\subsection{Metrics}

\begin{figure}[t]
    \centering
    \includegraphics[width=\linewidth]{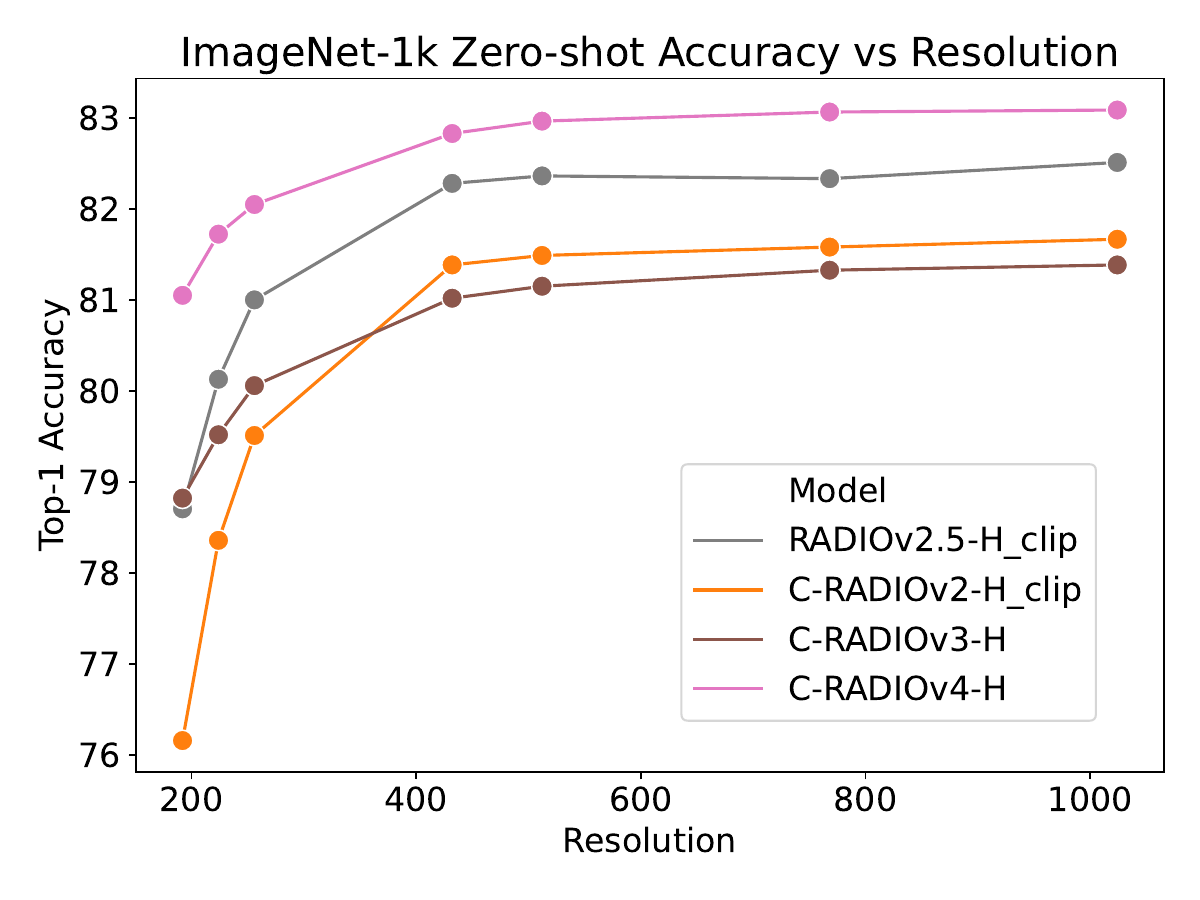}
    \caption{ImageNet-1K zero-shot accuracy as a function of input resolution.}
    \label{fig:zero_shot_resolution_scaling}
\end{figure}

In figure \ref{fig:zero_shot_resolution_scaling}, we show the zero-shot accuracy of RADIOv2.5 and various versions of C-RADIO. We note that RADIOv2.5 and C-RADIOv2 don't have a SigLIP2-aligned head, so we report the accuracy using their DFN CLIP head. In general, DFN CLIP is easier for the student model to match, and thus, up until C-RADIOv4, we were unable to match the zero-shot accuracy of our previous RADIOv2.5 model. All models shown exhibit strong resolution scaling, however, C-RADIOv4 strongly improves classification at low resolution relative to previous model generations. We achieve maximum zero-shot score at 1024px resolution, using aspect-preserving resizing.

\begin{figure}[t]
    \centering
    \includegraphics[width=\linewidth]{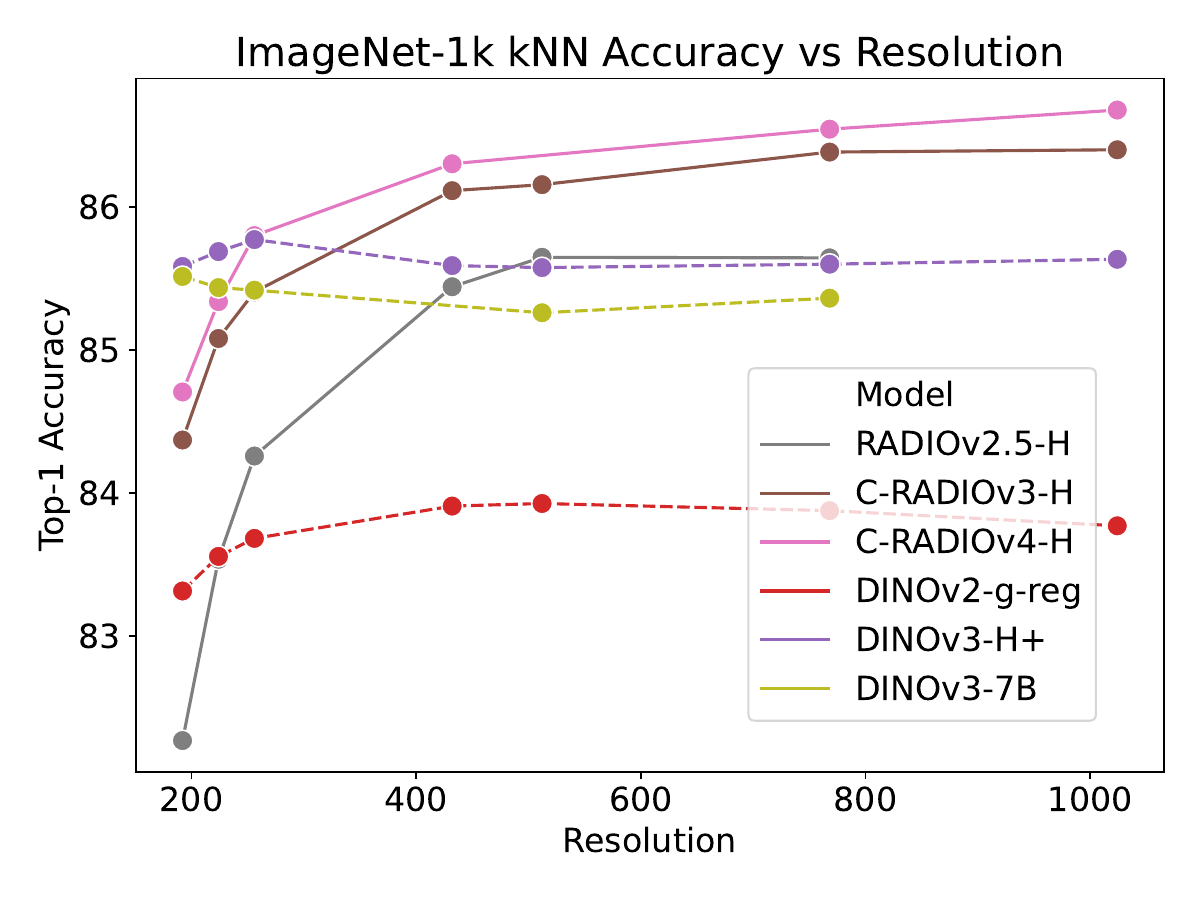}
    \caption{ImageNet-1K kNN accuracy as a function of input resolution.}
    \label{fig:knn_resolution_scaling}
\end{figure}

We're able to directly compare against DINOv2/3 models using k-NN (k-nearest neighbors) classification, as used by DINOv2 and prior RADIO works. We show these results in figure \ref{fig:knn_resolution_scaling}, where we can see DINOv3 has made large improvements on k-NN as compared to DINOv2, however, the ability to scale with resolution is concentrated around 192-256px, and higher resolutions degrade. For RADIO, the C-RADIO models drastically improve on RADIOv2.5, with C-RADIOv4 generally being a bit better than C-RADIOv3. This suggests that C-RADIOv3's issues in zero-shot more likely stem from issues matching SigLIP2 specifically, and not from lacking good holistic representations. Nonetheless, C-RADIOv4 is still an improvement for both tasks. Curiously, DINOv3's H+ model is better at kNN classification than the 7B model. Starting at 256px, C-RADIOv4-H is able to match or surpass DINOv3.

\begin{table}[]
    \centering
    \resizebox{\linewidth}{!}{
    \begin{tabular}{r|cccc}
        \multirow{2}{*}{Model} & \multirow{2}{*}{Depth} & Surface & \multirow{2}{*}{NAVI} & \multirow{2}{*}{SPair} \\
                               &                        & Normals &                       &                        \\
        \hline
        RADIOv2.5-H & 85.69 & 62.46 & 60.89 & 56.24 \\
        \hline
        C-RADIOv2-H & 85.02 & 60.10 & 59.82 & 53.98 \\
        C-RADIOv3-H & \bf{86.18} & \bf{62.52} & 62.10 & 58.54 \\
        \hline
        C-RADIOv4-SO440M & 85.29 & 61.91 & 62.44 & 60.01 \\
        C-RADIOv4-H & 85.55 & 61.70 & \bf{63.44} & \bf{60.57} \\
    \end{tabular}
    }
    \caption{Full Probe3d \cite{elbanani2024probing} evals for various RADIO models. Higher is better for all metrics.}
    \label{tab:probe3d_metrics}
\end{table}

\subsection{SAM3}

Older versions of RADIO had support for the SAM \cite{kirillov2023sam} adaptor, which allowed RADIO to replace SAM's vision encoder, while using their decoding and memory stack to perform the segmentation. This ability has found clever use in works such as RADSeg \cite{alama2025radsegunleashingparametercompute}, which uses the SAM head to refine segmentation masks and set a new state of the art for open vocabulary semantic segmentation. C-RADIOv4 upgrades our SAM support to SAM3 \cite{carion2025sam3segmentconcepts}, again allowing the core vision model to be replaced. We have a fork of the SAM3 codebase at \hyperlink{https://github.com/mranzinger/sam3-radio/tree/main}{https://github.com/mranzinger/sam3-radio} demonstrating how to do this replacement. In figures \ref{fig:sam3_basic_demo} and \ref{fig:sam3_query_demo}, we demonstrate RADIO's ability to replace the vision encoder, first with the image demo provided by SAM3, and second on our own image, with different queries. Qualitatively, RADIO has no issues acting in place of SAM3's Perception Encoder \cite{bolya2025perception} backbone. In table \ref{tab:saco_gold_instance} we show the benchmark results on SA-Co/Gold \cite{carion2025sam3segmentconcepts} instance segmentation. C-RADIOv4 becomes the second-best model, albeit, inheriting an uneven ability to replace SAM3's vision encoder. On ``metaclip\_nps'' and ``sa1b\_nps'', which are dominated by natural images, RADIO works quite well, with a smaller gap to SAM3, however, the gap becomes much larger on ``fg\_sports\_equipment'' and ``wiki\_common''. Improving on our ability to better match SAM3 is an open research direction.

\begin{figure}
    \centering
    \includegraphics[width=\linewidth]{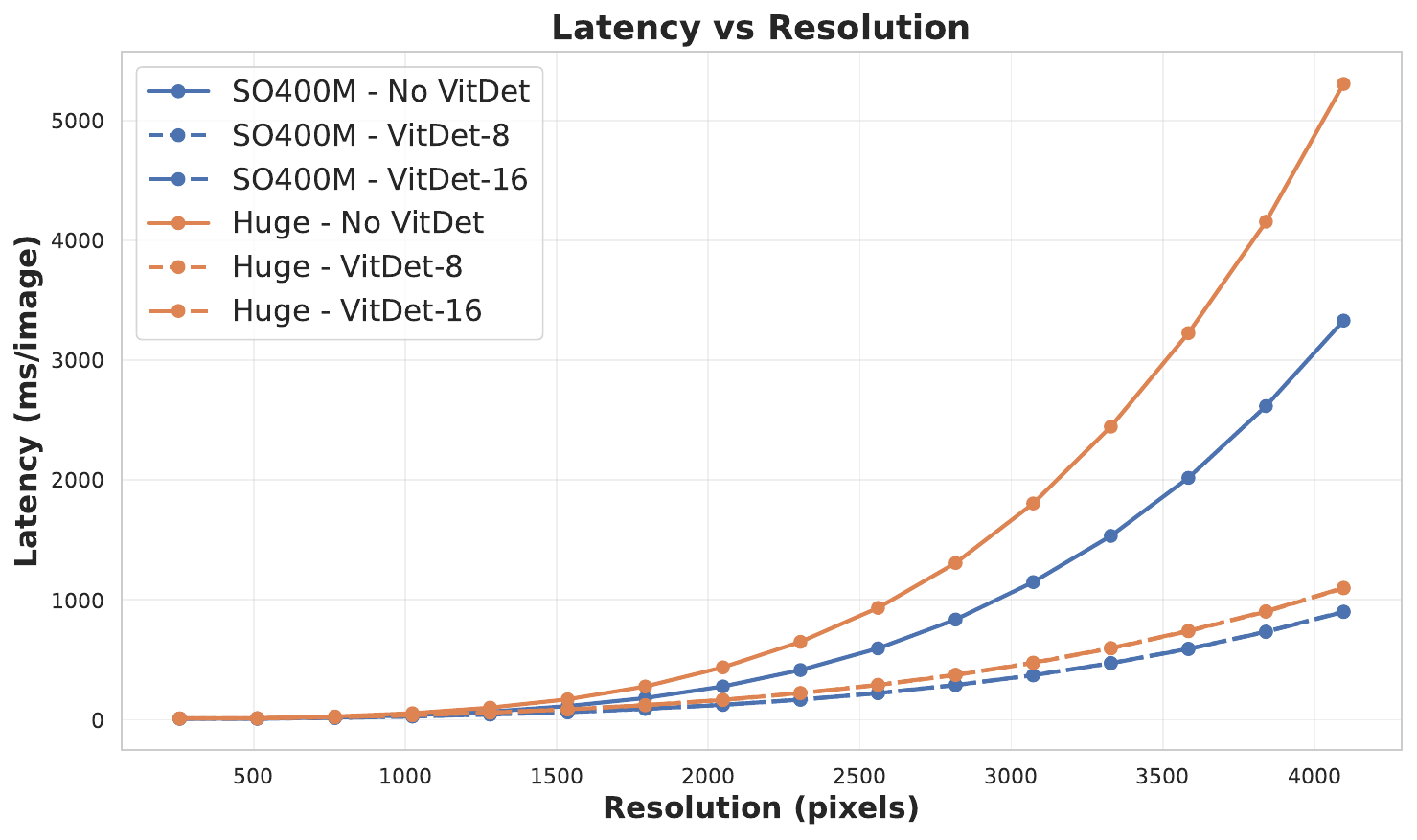}
    \caption{Latency analysis on A100 for both versions of C-RADIOv4, with and without ViTDet of a specified window size. The latency difference between ViTDet mode with window size 8 and 16 is negligible.}
    \label{fig:vitdet_latency}
\end{figure}

\begin{figure*}
    \centering
    \includegraphics[width=\linewidth]{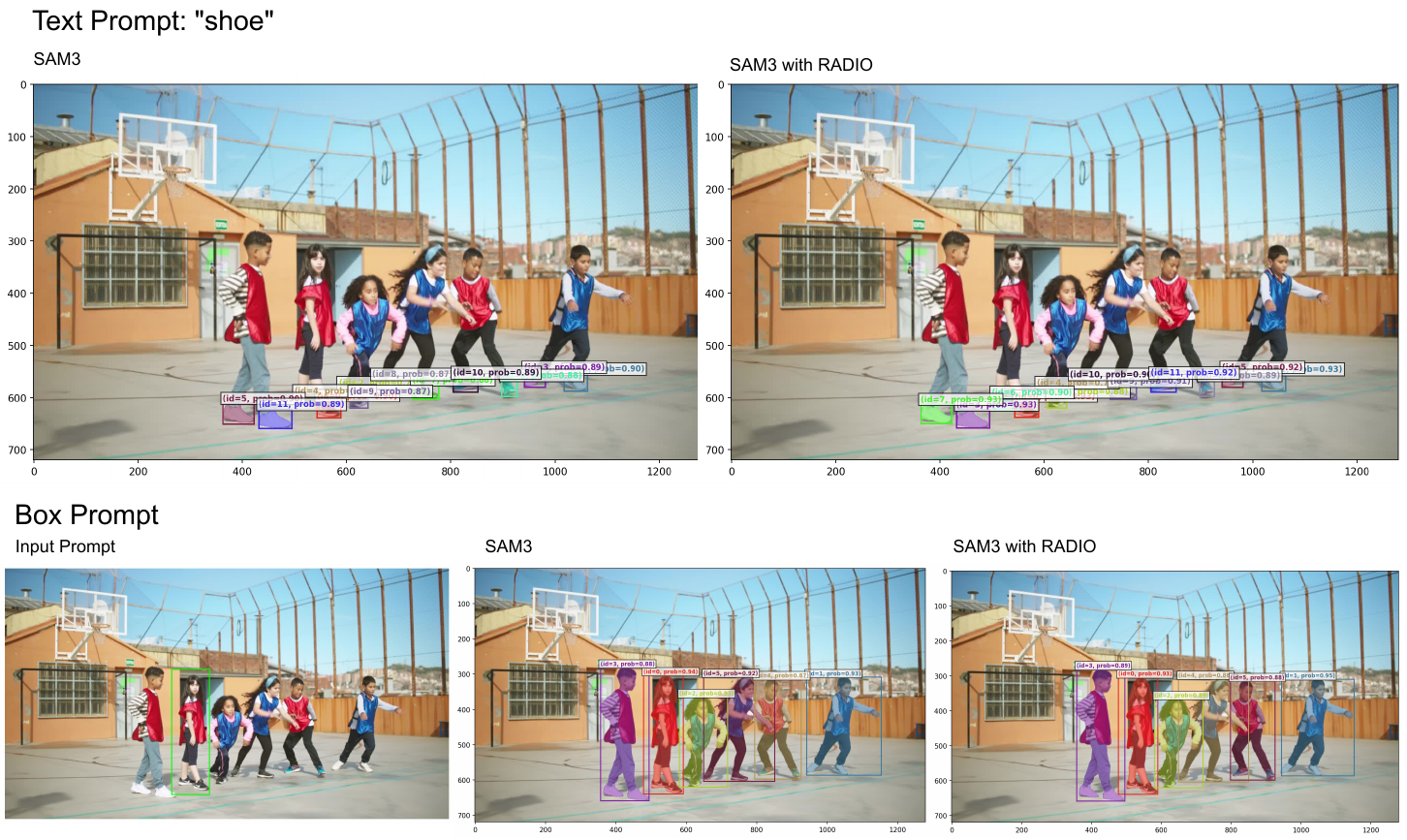}
    \caption{Mask results from the SAM3 image demo, either with regular SAM3, or with the vision encoder replaced by RADIO. RADIO is able to replicate the SAM3 results.}
    \label{fig:sam3_basic_demo}
\end{figure*}

\begin{figure*}
    \centering
    \includegraphics[width=\linewidth]{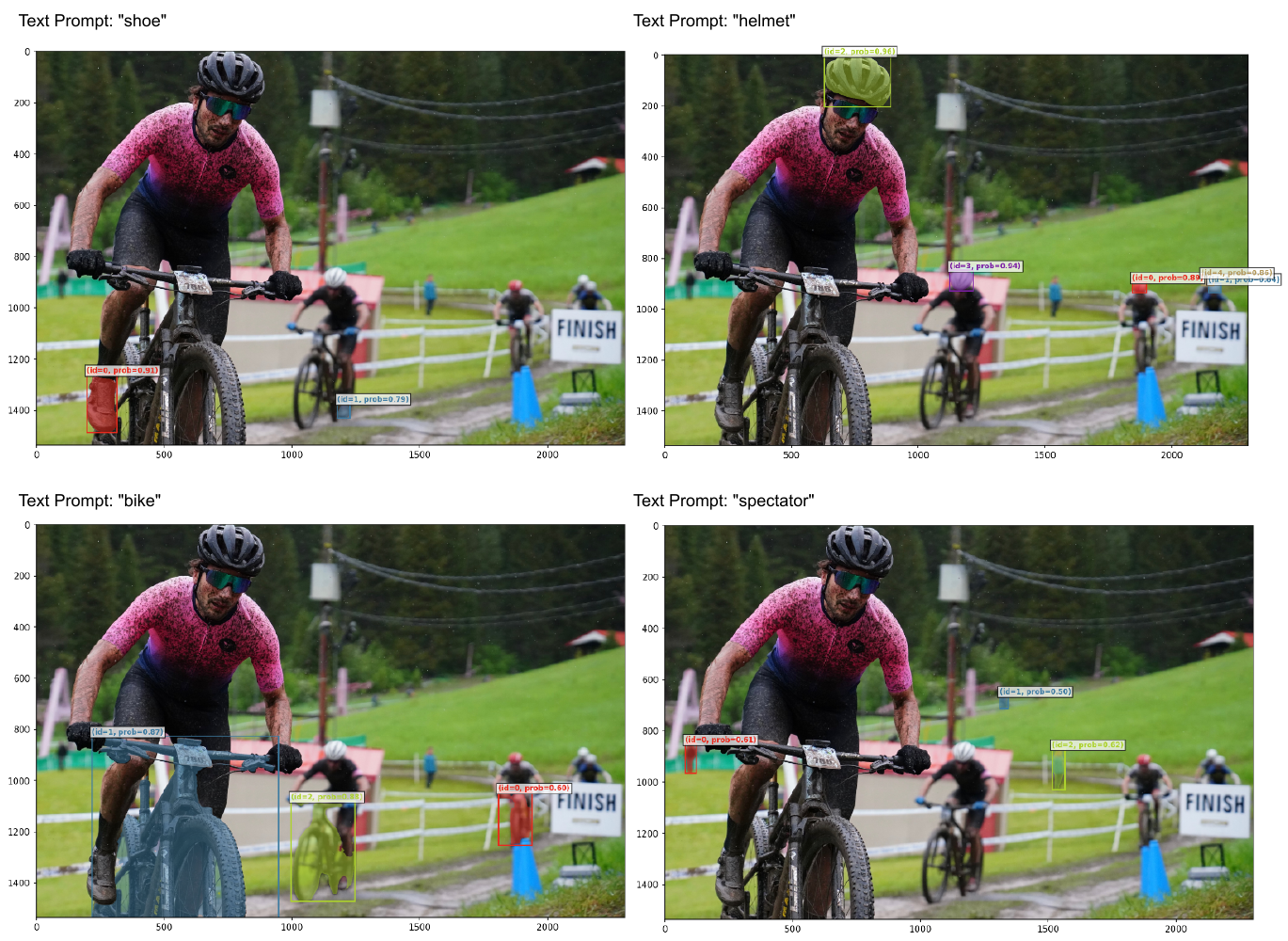}
    \caption{Text query results with RADIO replacing the SAM3 vision encoder, while keeping the decoder unchanged.}
    \label{fig:sam3_query_demo}
\end{figure*}

C-RADIOv4 supports "ViTDet-mode" \cite{li2022vitdet}, which allows the model to operate in either full global attention (e.g. ViTDet-mode disabled), or with mostly windowed attention, and a few global attention layers throughout. This can be controlled with the "vitdet\_window\_size" flag when constructing the model. SAM3 uses a ViT-L+ architecture, with 4 global layers, and the rest with windows of $24\times 24$ tokens. For increased efficiency, C-RADIOv4 supports windows anywhere between $6\times 6$ to $32 \times 32$ tokens, the only limitation being that the window size multiplied by the patch size needs to evenly divide the input image resolution. Smaller windows have the potential to be faster, as it reduces the quadratic penalty of self-attention \cite{vaswani2017attention}, however, hardware may reduce/eliminate the gap between similar window sizes, and smaller windows may come with a slight degradation in quality. We show the inference times for single-image inference on an A100 GPU in figure \ref{fig:sam3_benchmark_results}. For the SO400M model, a window size $\leq$ 12 is faster than SAM3's encoder. A window size of 8 for the ViT-H RADIO is nearly as fast. We also show the latencies for both C-RADIOv4 models for resolutions between 256px and 4096px, with and without ViTDet, in figure \ref{fig:vitdet_latency}. While ViTDet doesn't change the complexity of the model away from $O(T^2)$ with $T$ being the number of tokens, due to the fact that 4 layers still employ global attention, it does substantially reduce the growth factor. 

\paragraph{The curious case of ``person''}

\begin{figure*}
    \centering
    \includegraphics[width=\linewidth]{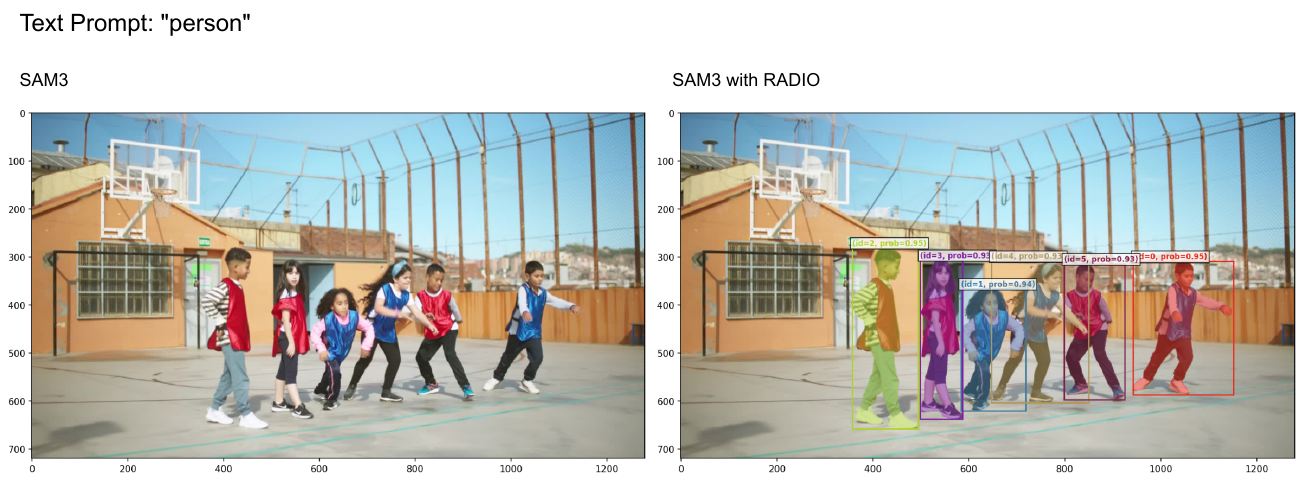}
    \caption{The SAM3 demo (\href{https://github.com/facebookresearch/sam3/issues/165}{github issue}) is unable to mask with the ``person'' query, while C-RADIOv4-[SO400M,H] works great.}
    \label{fig:sam3_person_failure}
\end{figure*}

\begin{figure*}
    \centering
    \includegraphics[width=\linewidth]{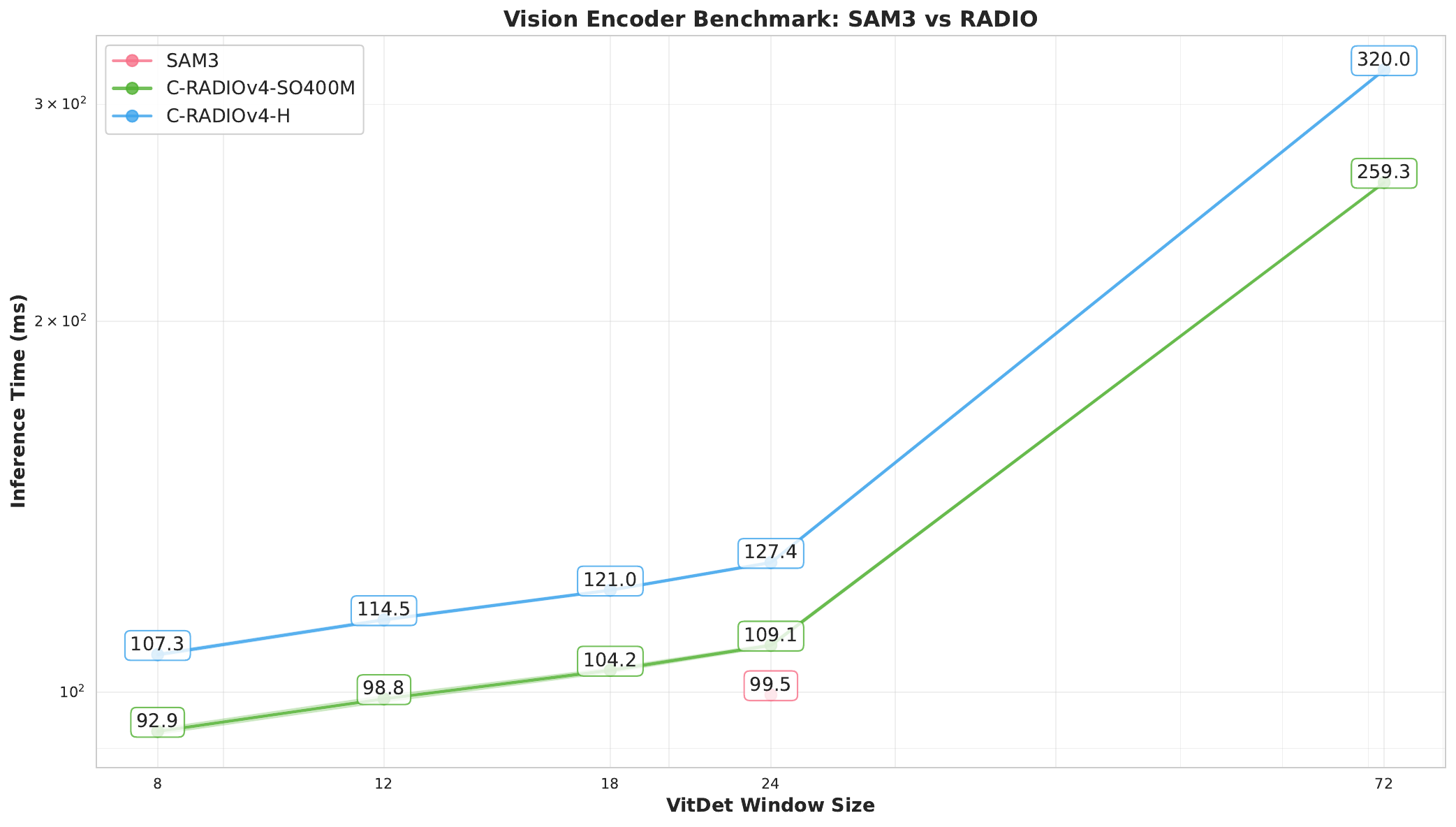}
    \caption{A100 single-image benchmarking results for the vision encoder of SAM3 vs RADIO, for both SO400M and Huge sizes. SAM3 uses a ViT-L+ architecture with window size 24. A window size of 72 is equivalent to full global attention.}
    \label{fig:sam3_benchmark_results}
\end{figure*}

On the SAM3 github (as of 1/14/2026), there is the github \href{https://github.com/facebookresearch/sam3/issues/253}{issue 253} that points out that the github example for SAM3 doesn't work properly with the ``person'' query. We are able to replicate this behavior. However, with C-RADIOv4 acting as the vision backbone, the query works as expected. We show this behavior in figure \ref{fig:sam3_person_failure}. The RADIO variant works in global attention mode, as well as with ViTDet with window size 8 (others would probably work too, these are the only two we tested). This seems to suggest that the slight difference in representations between the two vision encoders is having a thresholding effect with this particular query.

\begin{table*}[]
    \centering
    \resizebox{\linewidth}{!}{
    \begin{tabular}{r r|ccccccc|c}
        & \multirow{2}{*}{Model} & \multicolumn{8}{c}{SA-Co/Gold Instance Segmentation (cgF1)} \\
        &                       & metaclip\_nps & sa1b\_nps & crowded & fg\_food & fg\_sports\_equipment & attributes & wiki\_common & Avg \\
        \hline
        & Human      &               &           &         &          &                       &            &               & 72.8 \\
        & OWLv2      &               &           &         &          &                       &            &               & 24.6 \\
        & DINO-X     &               &           &         &          &                       &            &               & 21.3 \\
        & Gemini 2.5 &               &           &         &          &                       &            &               & 13.0 \\
        \hline
        & SAM3       & 47.3          & 53.7      & 61.1    & 53.4     & 65.5                  & 54.9       & 42.5          & 54.1 \\
        \hline
        \multirow{4}{*}{\rotatebox[origin=c]{90}{C-RADIOv4\ }}
        & SO400M-VDT8 & 43.0         & 44.5      & 54.9    & 38.4     & 38.4                  & 40.3       & 22.2          & 40.3 \\
        & SO400M-G    & 43.8         & 45.7      & 55.9    & 40.1     & 39.8                  & 41.6       & 23.1          & 41.4 \\
        & H-VDT8      & 45.2         & 48.1      & 56.6    & 40.3     & 45.3                  & 44.0       & 26.2          & 43.7 \\
        & H-VDT12     & 45.6         & 48.4      & 57.3    & 40.2     & 46.1                  & 45.2       & 26.7          & 44.2 \\
        & H-G         & 45.9         & 48.8      & 57.4    & 40.9     & 46.5                  & 45.9       & 27.3          & 44.7 \\
    \end{tabular}
    }
    \caption{Results on SA-Co/Gold \cite{carion2025sam3segmentconcepts} instance segmentation. ``VDT'' refers to ViTDet window size $W$, while ``G'' refers to global attention throughout.}
    \label{tab:saco_gold_instance}
\end{table*} 
\section{Conclusion}\label{sec:conclusion}

Owing to improvements in base foundation models, particularly DINOv3 \cite{simeoni2025dinov3}, as well as improvements to our distillation algorithm, C-RADIOv4 enjoys large improvements over its predecessors. Different from previous releases, we include an SO400M \cite{alabdulmohsin2023so400m} version, which is often able to be competitive with ViT-H, while being cheaper. We demonstrate how C-RADIOv4 may be used to replace the vision encoder in SAM3 \cite{carion2025sam3segmentconcepts}, and with the SO400M variant with ViTDet window size $\leq 12$, it's even faster than the ViT-L+ Perception Encoder \cite{bolya2025perception} that SAM3 uses. Past RADIO models have seen widespread usage, including in the Nemotron Nano V2 VL \cite{nvidia2025nvidianemotronnanov2} vision-language model, autonomous vehicles, robotics, OCR document parsing \cite{chumachenko2025nvidianemotronparse11}, open vocabulary semantic segmentation \cite{alama2025radsegunleashingparametercompute,alama2025rayfronts}, and many more. Given the commercially permissive license, we hope that both the academic and industrial community will be able to leverage this foundational model to build great things.

{
    \small
    \bibliographystyle{ieeenat_fullname}
    \bibliography{main}
}

\end{document}